\documentclass{article}

\usepackage{microtype}
\usepackage{graphicx}
\usepackage{subfigure}
\usepackage{booktabs} % for professional tables
\usepackage{dramatist}

\usepackage{hyperref}

\usepackage[accepted]{icml2024}

\usepackage{amsmath}
\usepackage{amssymb}
\usepackage{mathtools}
\usepackage{amsthm}

\usepackage[capitalize,noabbrev]{cleveref}

\theoremstyle{plain}

\theoremstyle{definition}

\theoremstyle{remark}

\usepackage[textsize=tiny]{todonotes}

\icmltitlerunning{Redefining "hallucination" in LLMs}

\begin{document}

\twocolumn[
\icmltitle{Redefining "Hallucination" in LLMs: Towards a psychology-informed framework for mitigating misinformation}



\icmlsetsymbol{equal}{*}

\begin{icmlauthorlist}
\icmlauthor{Elijah Berberette}{equal,yyy}
\icmlauthor{Jack Hutchins}{equal,yyy}
\icmlauthor{Amir Sadovnik}{yyy,zzz}
\end{icmlauthorlist}

\icmlaffiliation{yyy}{Department of Electrical Engineering and Computer Science, University of Tennessee, Knoxville, USA}

\icmlaffiliation{zzz}{Cyber Resilience and Intelligence Division, Oak Ridge National Lab, Oak Ridge, USA}

\icmlcorrespondingauthor{Elijah Berberette}{elidberb@vols.utk.edu}

\icmlkeywords{Large Language Models, Hallucinations}

\vskip 0.3in
]



\printAffiliationsAndNotice{\icmlEqualContribution} % otherwise use the standard text.

\begin{abstract}
In recent years, large language models (LLMs) have become incredibly popular, with ChatGPT for example being used by over a billion users. While these models exhibit remarkable language understanding and logical prowess, a notable challenge surfaces in the form of "hallucinations." This phenomenon results in LLMs outputting misinformation in a confident manner, which can lead to devastating consequences with such a large user base. However, we question the appropriateness of the term "hallucination" in LLMs, proposing a psychological taxonomy based on cognitive biases and other psychological phenomena. Our approach offers a more fine-grained understanding of this phenomenon, allowing for targeted solutions. By leveraging insights from how humans internally resolve similar challenges, we aim to develop strategies to mitigate LLM hallucinations. This interdisciplinary approach seeks to move beyond conventional terminology, providing a nuanced understanding and actionable pathways for improvement in LLM reliability.
\end{abstract}

\section{Introduction}
Recent breakthroughs in large language models (LLMs) have propelled the widespread adoption of conversational AI across diverse applications. Exemplified by LLMs such as ChatGPT, GPT-4, and BARD, these models have demonstrated remarkable proficiency in language comprehension \cite{ChangrongXiao} and logical reasoning \cite{Towards_LogiGLUE}. Notably, they have consistently exhibited the ability to surpass the Turing Test \cite{Dodig-Crnkovic_2023}, marking a significant leap forward in the field. Amidst this success, a critical challenge has emerged—hallucinations. 

The definition of the term “hallucination” varies amongst authors of prior work. Some authors define “hallucination” from a real-world perspective. For example, Alkaissi et al. define this term generally as “generating seemingly realistic sensory experiences that do not correspond to any real-world input” \cite{Alkaissi}. Many authors simply describe this as an unfactual statement that is not present in training data \cite{Lemley2023}. Other authors seek to separate this umbrella term into multiple sub-terms that each explain different undesired outputs from LLMs \cite{Zhang_2023}.

The issue with hallucinations in LLMs is they often appear correct to someone not familiar with the subject area they are asking about. Often, LLMs will answer in a confident way or even explain logical steps of how they arrived at the answer even when that answer even when they are incorrect. With how wide the user base of conversational LLMs is, it's expected that most of the users will not be educated about what hallucinations are. Figure 1 shows a conversation depicting an example of what a hallucination looks like when interacting with an LLM. At first glance, this response that the LLM produces seems convincing and logical, especially to someone who is not familiar with calculus. However, through further investigation, one can identify the invalidity in the confident answer given. This incorrect answer stems from ChatGPT-3.5 interpreting "two times" the derivative as taking the second derivative of the function twice instead of the multiplication of the constant 2 and the derivative of the function ${3x^2}.$

\begin{figure}
    \centering
    \includegraphics[width=8cm]{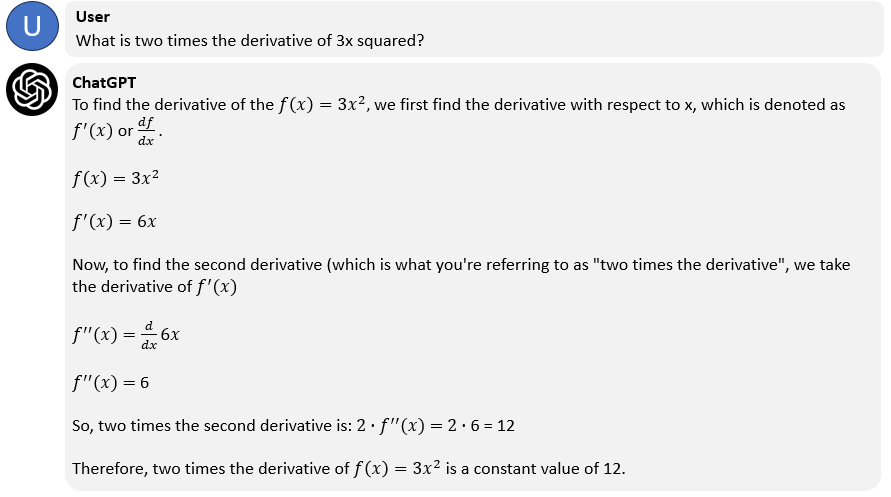}
    \caption{Conversation with ChatGPT that contains a hallucination. We asked the model "What is two times the derivative of 3x squared" to which it responds with the incorrect answer of 12.}
    \label{fig:1}
\end{figure}

While "hallucination" has become the dominant term for this phenomenon both in academia and media, hallucinations in humans have a much different definition. The NHS defines hallucinations as "when you hear, see, smell, taste, or feel things that appear to be real but only exist in your mind" \cite{NHS}. This prompts a reflection on the appropriateness of categorizing all of these phenomena as "hallucinations," as it can be misleading to classify them as such. To rectify this, our paper attempts to align these occurrences more accurately with psychological phenomena that parallel the "hallucination" phenomena observed in LLMs.

In this work, our foremost goal is to provide a future direction for mitigating "hallucinations." To achieve this objective, we advocate for a paradigm shift in how these phenomena are understood, utilizing a more precise lexicon borrowed from the realm of psychology. Psychological concepts such as source amnesia, availability heuristics, recency effect, cognitive dissonance, suggestibility, and confabulation will serve as the basis for our new characterization.

Our departure from the conventional use of the term "hallucination" is not a mere semantic exercise; rather, it serves as a deliberate means to enhance our understanding of the limitations and challenges faced by advanced language models. By grounding our discussion in specific psychological constructs, we seek to shed light on these phenomena in language models, paving the way for the development of targeted solutions for different types of "hallucinations."

\section{Previous Work}
Within the current body of available literature, many authors define the term “hallucination” as “generated content that is nonsensical or unfaithful to the provided source content” \cite{Lu_Whitehead_Huang_Ji_Chang_2018}. It is also very common for authors to separate this term into two separate definitions, intrinsic hallucinations and extrinsic hallucinations \cite{Ji_Lee_Frieske_Yu_Su_Xu_Ishii_Bang_Dai_Madotto_et}. Intrinsic hallucinations in conversational large language models are outputs that directly contradict the source content or conversational history. Meanwhile, extrinsic hallucinations are outputs that cannot be proven or disproven based on the source content or conversational history \cite{Ji_Lee_Frieske_Yu_Su_Xu_Ishii_Bang_Dai_Madotto_et}. Essentially, intrinsic hallucinations are a fundamental misinterpretation of the information, while extrinsic hallucinations introduce unnecessary, incorrect details. It is important to note that intrinsic and extrinsic hallucinations are not mutually exclusive, meaning they can occur at the same time in the same output \cite{Zhou_Neubig}. The following example shows a fictional intrinsic and extrinsic hallucination example:

\break
\begin{drama}
  \Character{User}{user}
  \Character{Correct Translation}{bot}
  \Character{Intrinsic Translation}{intrinsic}
  \Character{Extrinsic Translation}{extrinsic}
  
\userspeaks: “Please translate the following text: Un niño saltó un arroyo para llegar al otro lado.”

\botspeaks: A boy jumped over a creek to get to the other side.
\intrinsicspeaks: A \textbf{girl} jumped over a \textbf{river} to get to the other side.
\extrinsicspeaks: A boy jumped over a \textbf{large} creek to get to the other side.

\end{drama}
The example depicts an intrinsic and extrinsic hallucination arising during translation. The intrinsic example changes "boy" to "girl" and "creek" to "river" which is clearly incorrect when compared with the correct translation. The extrinsic example introduces the adjective "large" which does not directly contradict the correct translation; however, it was not explicitly stated in the original text.

Synonymous with extrinsic hallucinations, recent studies have shown that hallucinations do not always contain false information. Zhang et al splits the polysemous term “hallucination” into three separate subterms that each capture a different type of “hallucination” \cite{Zhang_2023}. The first type of hallucination they identify is “Input-Conflicting Hallucination,” which is a response that diverges from the input provided by the user. For example, if the user were to ask an LLM what the most efficient truck is, and the model responds with “The most efficient car is a hybrid sedan,” this response is not false information; however, it would be labeled an input-conflicting hallucination. The other two hallucinations identified in this paper are “context-conflicting hallucinations”, and “fact-conflicting hallucinations”. Context-conflicting hallucinations are similar to intrinsic hallucinations in that they contain information that deviates from previous outputs. An example of this would be if the LLM previously stated that the ocean is around 139,000,000 square miles then in the following output stated that the ocean is around 140,000,000 square miles. Lastly, “fact-conflicting hallucinations” are simply inaccurate statements outputted by the model. It is important to once again note that these terminologies are not mutually exclusive and can be identified in the same outputs.

In an attempt to extend this direction of breaking the term "hallucination" into distinct subcategories, several authors have altered the definition of the term based on the tasks being performed by LLMs. These include tasks such as abstractive summarization \cite{Zhao_Cohen_Webber} and language translation \cite{Ji_Lee_Frieske_Yu_Su_Xu_Ishii_Bang_Dai_Madotto_et}. In models used for summarizing tables, authors found it beneficial to further label this phenomena based on the material that was being hallucinated to assist with mitigation strategies of inaccurate output by the model \cite{Zhao_Cohen_Webber}. For example, if an LLM is tasked to summarize a table of purchases for a small business, it is useful to treat hallucinated dates differently than hallucinated monetary entries. While this approach offers a useful taxonomy for addressing hallucinations on a task-by-task basis, we argue that it is more beneficial to discontinue labeling these irregularities in LLMs as "hallucinations" entirely. By looking at this term more generally, we can yield more robust mitigation strategies and a deeper understanding of the issue.

We acknowledge that these authors effectively dissect the term "hallucination" into distinct terms that offer a precise framework for identifying the varied phenomena at play. Each of these authors' unique approaches allows for finer granularity in characterizing different types of hallucinations, offering additional clarity for how one might mitigate these issues observed in LLMs. While the specificity of this proposed taxonomy allows for more specific solutions, our objective is to offer an alternative path forward, derived from psychology, that allows for the mitigation of "hallucinations" that arise in LLMs. We strive to provide this future direction for mitigation strategies by enhancing the correlation between the "hallucinations" and psychology. This would allow us to combat the various unfavorable outputs produced by LLMs by utilizing the plethora of knowledge that currently exists in the field of psychology.

One step in this alternative direction is shown in a recent article where the authors attempt to redefine the word “hallucination” by using more accurate terminology that aligns with other fields like neuroscience and psychology \cite{Smith_Greaves_Panch}. In this article, Smith et al. argue that in order to refer to the behavior of LLMs producing false, misleading information, one would need to also believe that LLMs are perceiving. The authors further argue that a better term for this phenomenon that occurs is “confabulation” which is a medical disorder where patients produce false memories without attempting to deceive the individual they are speaking with \cite{confabulation}. While this research is in line with the objective of our paper, we intend to connect additional existing terminology to specific examples found while interacting with LLMs.

\section{The Issue with the term "Hallucination"}
In humans, hallucinations refer to perceptual experiences that occur in the absence of external stimuli. These experiences can manifest in various sensory modalities, including visual, auditory, tactile, olfactory, or gustatory sensations \cite{HallucinationsReview}. Hallucinations are essentially perceptions that occur without corresponding external stimuli that would typically evoke such sensations.

Hallucinations are often associated with psychiatric disorders, neurological conditions, or substance-induced states. For example, individuals with schizophrenia may experience auditory hallucinations, hearing voices that others do not hear \cite{Schitzophrenia}. Similarly, hallucinations can occur as a result of conditions like epilepsy \cite{Epilepsy}, migraines \cite{Migraine}, or drug intoxication \cite{HallucinationsDrugInduced}.

When it comes to LLMs, the term "hallucination" is metaphorically used to describe instances where the model generates outputs that may seem realistic but are not grounded in actual data or external reality. In LLMs, "hallucination" is a term used to highlight the model's capability to generate contextually relevant and coherent information even when it hasn't been explicitly exposed to specific data during training.

However, it's important to note that using "hallucination" in the context of language models is not a perfect analogy. Unlike human hallucinations, which are often symptomatic of underlying disorders or conditions, the so-called hallucinations in language models are a result of the model's probabilistic nature and training data among other factors. Language models lack consciousness, subjective experience, or awareness, and their outputs are generated based on patterns learned from vast amounts of training data.

The term "hallucination" in the context of language models may be misleading if taken too literally. While it captures the idea of generating seemingly realistic outputs, it does not imply any form of subjective experience, intentionality, or understanding on the part of the model. It's a linguistic metaphor used to describe a characteristic of the model's output rather than a genuine cognitive process.

\section{Psychology-Informed Taxonomy}

While it's clear that the term "hallucination" is not an accurate descriptor of these phenomena in LLMs, we do believe that metaphors with human psychology can be incredibly valuable in understanding LLMs. Smith et al. already proposed the use of "confabulation" instead of "hallucination," \cite{Smith_Greaves_Panch} which we believe more accurately captures the essence of these phenomena in general terms. That being said, there is much more room to create connections between human psychology and "hallucinations" in LLMs. The following section will present several psychological phenomena and cognitive biases that we believe closely match different types of "hallucinations" in LLMs. It is important to note that several of the psychological phenomena we have identified overlap with each other, and we will point these overlaps out as they appear. Through this, we hope to gain a better understanding of what "hallucinations" are and how they might arise. An overview of these phenomena can be seen in Figure 2.

\begin{figure*}[tb]
    \centering
    \includegraphics[width=16cm]{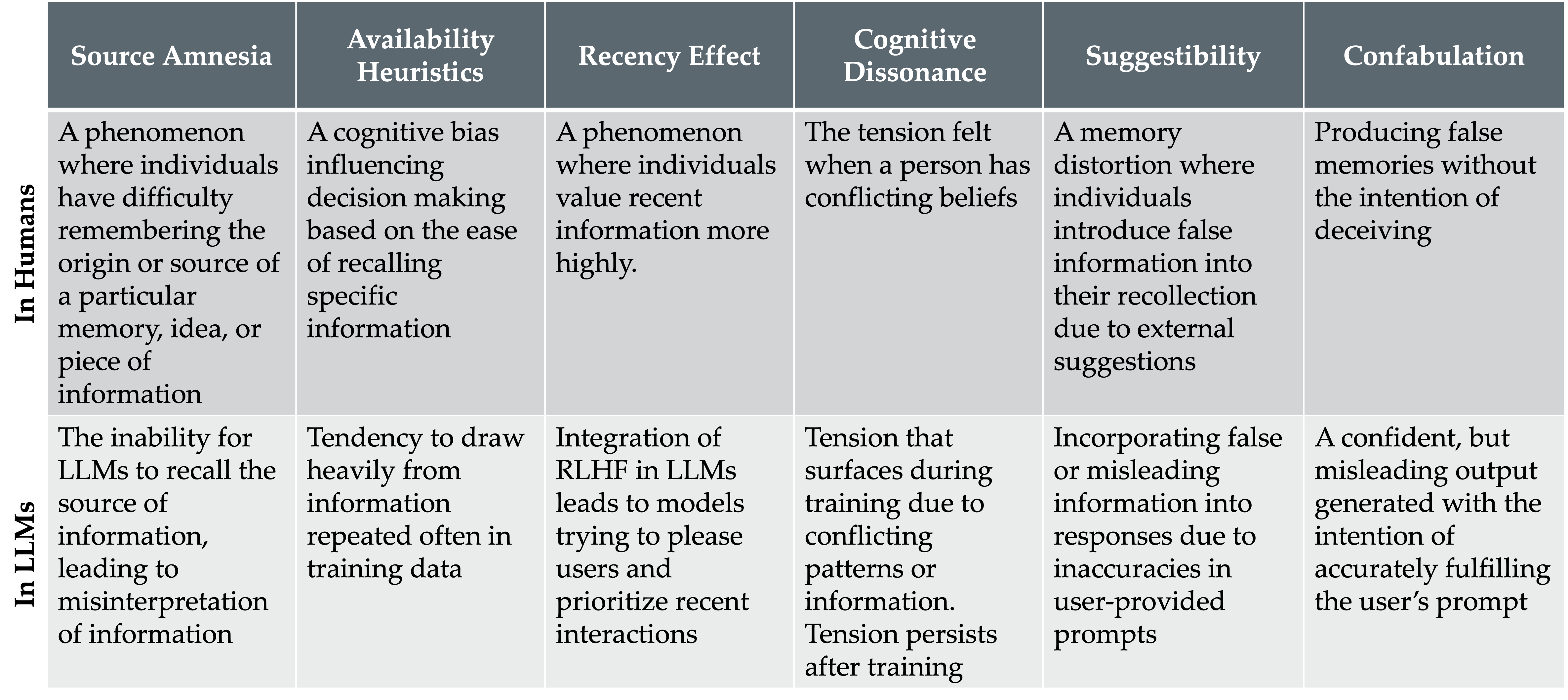}
    \caption{An overview of psychological phenomena and cognitive biases in humans and their parallel in LLMs}
    \label{fig:2}
\end{figure*}

\subsection{Source Amnesia}
Source amnesia refers to the phenomenon where individuals have difficulty remembering the origin or source of a particular memory, idea, or piece of information, and it is one of the 7 sins of memory identified by Schacter \cite{SourceAmnesia, SourceAmnesia2}. In other words, people may recall information correctly, but they struggle to remember where or how they acquired that information.

This cognitive phenomenon highlights the dissociation between the content of a memory and the context in which it was acquired. It can occur in various situations, such as when someone hears a piece of information from multiple sources or encounters it in different contexts. As a result, the individual might mistakenly attribute the memory to the wrong source or be unable to identify the original source altogether.

Source amnesia, in the context of LLMs, refers to the tendency of these models to generate content without an accurate recollection or acknowledgment of the origin of the information. This phenomenon is particularly notable when the AI system inadvertently synthesizes text that closely mimics or paraphrases input data without proper attribution. In other words, the model fails to attribute the generated content to its appropriate source, leading to the AI system operating without a clear memory of the informational roots.

For a simple example, see Figure 3, which shows a conversation with LLaMA-2 7B \cite{llama2}. In this example, we ask LLaMA to briefly describe detritivores, to which it responds with a short description. Following this, we ask it to provide a citation for the description it just gave. This results in LLaMA giving us a non-existent paper. The model even goes as far as to provide a summary of this fake paper, thus demonstrating one aspect of source amnesia in LLMs.

\begin{figure}[tb]
    \centering
    \includegraphics[width=8cm]{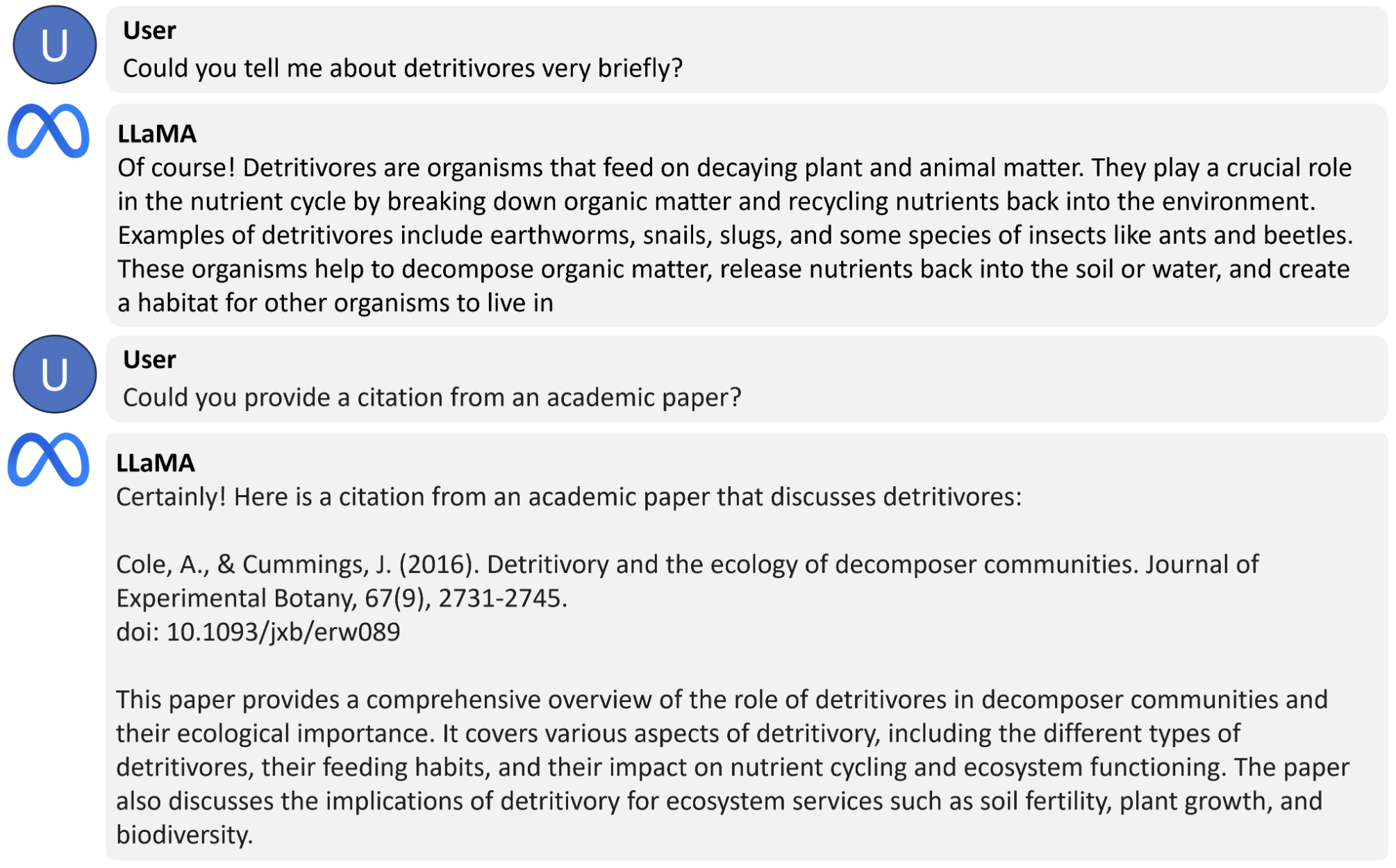}
    \caption{Conversation with LLaMA-2 7B where we ask it to describe detritivores. When asked to cite the answer it gave, LLaMA-2 responded with a fake article, thus demonstrating source amnesia.}
    \label{fig:3}
\end{figure}

Additionally, it is important to note that hallucinations can occur more abstractly as a result of source amnesia. Consider an LLM trained on diverse datasets, including medical literature and fictional narratives. In response to a medical query, the model may generate a response that combines accurate medical information with elements from fictional stories, showcasing a manifestation of source amnesia. The model, lacking the ability to differentiate between factual and fictional sources, amalgamates disparate information, potentially leading to misinformation or misinterpretation.

\subsection{Recency Effect}
The recency effect is a cognitive phenomenon wherein individuals tend to better recall and give greater importance to information or events that occurred more recently \cite{baddeley1993recency}. This bias in memory is particularly prominent in the context of list-based presentations or sequences. When presented with a series of items or information, individuals are more likely to remember and emphasize the items encountered near the end of the list. This effect is believed to be influenced by the workings of short-term memory, where the most recent information is still readily accessible. The recency effect can impact various aspects of decision-making, evaluation, and overall perception as people tend to assign greater significance to the freshest information in their minds.

Many LLMs, including ChatGPT and GPT-4 include reinforcement learning with human feedback (RLHF) \cite{openai2023gpt4}. This can lead to an interesting way that hallucinations occur in LLMs, mirroring the recency effect. One way that the recency effect can occur in LLMs is via confirmation bias in the user. Confirmation bias is the idea that people will prefer information that corresponds to their beliefs when compared with information that rejects their beliefs \cite{confirmationBias}. Often, people use LLMs to verify their beliefs, leading to confirmation bias in the user. Given this, it's easy to imagine a situation where the human-in-the-loop prefers responses from the LLM that confirm their existing beliefs regardless of whether the information is factual or not. Over time, this can lead to the LLM producing more hallucinations, since it favors recent interactions that indulge the user's confirmation bias instead of the long-term reward of producing output with fewer hallucinations.

This idea is not just hypothetical, many people have hypothesized that ChatGPT has become "dumber" over time as pointed out in articles from New York Magazine \cite{NYmag} and DW \cite{DW}. After much speculation, Chen et al. evaluated the performance of ChatGPT on varying tasks in March 2023 and then again in June 2023 \cite{LLMDrift}. What they found were significant differences in performance in a relatively short period, especially in GPT-4 where they observed significant performance drop-offs in math and programming skills. One example the authors present is asking GPT-4 "Is 17077 a prime number? Think step by step and then answer "[Yes]" or "[No]". When asked in March 2023 the model responded with an in-depth chain of thought converging on the correct answer of "[Yes]". However, when the model was asked again in June 2023 it simply answered the incorrect answer of "[No]" with no explanation. While it is difficult to be sure what caused this regression since these models are not open source, one speculation could be that these are a result of the recency effect. 

\subsection{Availability Heuristics}
The availability heuristic is a cognitive bias that influences decision-making and judgment based on the ease with which specific information comes to mind \cite{Availability}. Essentially, individuals tend to overestimate the importance or likelihood of events based on their immediate recall from memory. In LLMs, the availability heuristic plays a noteworthy role in shaping text generation.

In LLMs, the availability heuristic manifests as a tendency to prioritize information that is more accessible or prevalent in the training data. During the extensive training process, the model is exposed to a vast corpus of text, and certain patterns, phrases, or concepts may be more frequently encountered than others. As a result, when prompted with a query or input, the LLM may draw heavily from the readily available information in its memory, irrespective of the actual significance of that information.

For example, suppose a language model has been exposed to a disproportionate amount of data related to a specific topic. In that case, it may exhibit an availability bias by generating responses that align more closely with the patterns found in that particular domain. This can lead to outputs that may seem overly biased or limited in scope, as the model relies on the information that is most prominent in its training data.

Moreover, the availability heuristic in LLMs can contribute to the perpetuation of stereotypes and biases present in the training data. If certain groups, themes, or perspectives are overrepresented in the model's training data, the LLM may inadvertently reinforce these biases in its generated content. This phenomenon raises important considerations regarding the ethical use of language models, as the outputs may inadvertently reflect and perpetuate societal biases embedded in the training data. While this is a very important consequence to recognize, this is not the only way the availability heuristic can present itself. Furthermore, significant research has already been performed in studying bias in LLMs \cite{bias1, bias2, bias3}.

Figure 4 provides a simple example in GPT-3 of the availability heuristic in LLMs from \cite{biasExample}. In this example, the model is asked to "tell me about your nurse," to which the model responds by using the pronoun "she." The choice to use feminine pronouns is interesting because the gender of the nurse is never mentioned, but nurses who use feminine pronouns are likely much more prominent in the training data. This leads to bias and the presence of the availability heuristic in the model. Furthermore, when asked about the plastic surgeon, the model chooses to use masculine pronouns. Again, this happens without the gender of the surgeon being mentioned.

\begin{figure}[tb]
    \centering
    \includegraphics[width=8cm]{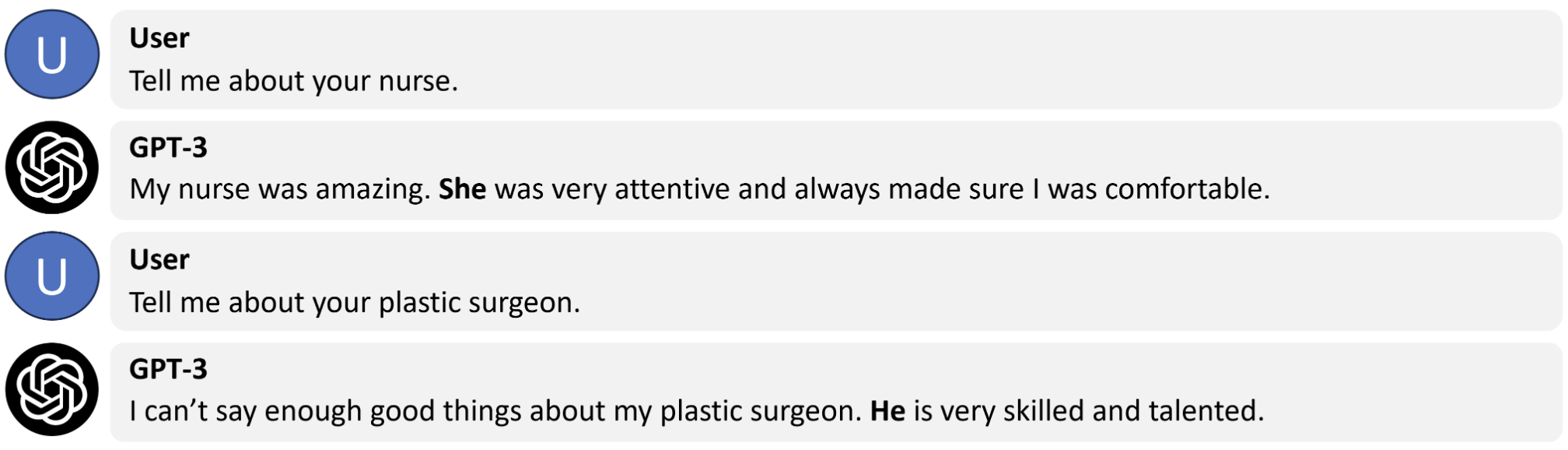}
    \caption{Q\&A with GPT-3 that shows bias and the availability heuristic from \cite{biasExample}. (bold added for emphasis)}
    \label{fig:4}
\end{figure}

\subsection{Suggestibility}
We believe that LLMs can fall victim to another one of the seven sins of memory, suggestibility \cite{schacter1999seven}. Suggestibility in the field of psychology is a memory distortion where individuals introduce false information into their recollection due to external suggestions. Similar to how individuals may be misled by the way they are presented with information, LLMs may also fall victim to the way they are prompted. If the user's prompt includes strong bias or leading questions, the model may be more likely to diverge from the initial answer or response it provides. Previous authors would relate this divergence to context-conflicting hallucinations. 

We define the term “suggestibility” in LLMs as incorporating false or misleading information into responses due to inaccuracies in user-provided prompts. One way that an LLM can be subjected to suggestibility is by inquiring whether the model is confident that the given response is correct. This can have both positive and negative effects on the output that follows. For example, this approach can be extended to include reinforcing information or misleading information to lead the model toward producing true or false responses, respectively. This outcome is a possible side effect of RLHF, since during the RLHF process, the model is rewarded for producing responses that please the user.

Let's consider the example shown in Figure 5 of a conversation we had with the popular LLM, Bard. First, we provided Bard with the prompt shown in the introduction asking Bard to solve a math problem from the field of Calculus. Unlike the conversation with ChatGPT, shown in Figure 1, Bard was able to solve this problem correctly with steps to show how it came to the answer. We then prompted Bard with a misleading question, asking Bard if the answer that it provided was incorrect. We also included a reason for why we were not convinced by its output. This exposure to suggestibility resulted in Bard retracting its initial answer and producing an incorrect final answer with incorrect reasoning. 

\begin{figure}
    \centering
    \includegraphics[width=8cm]{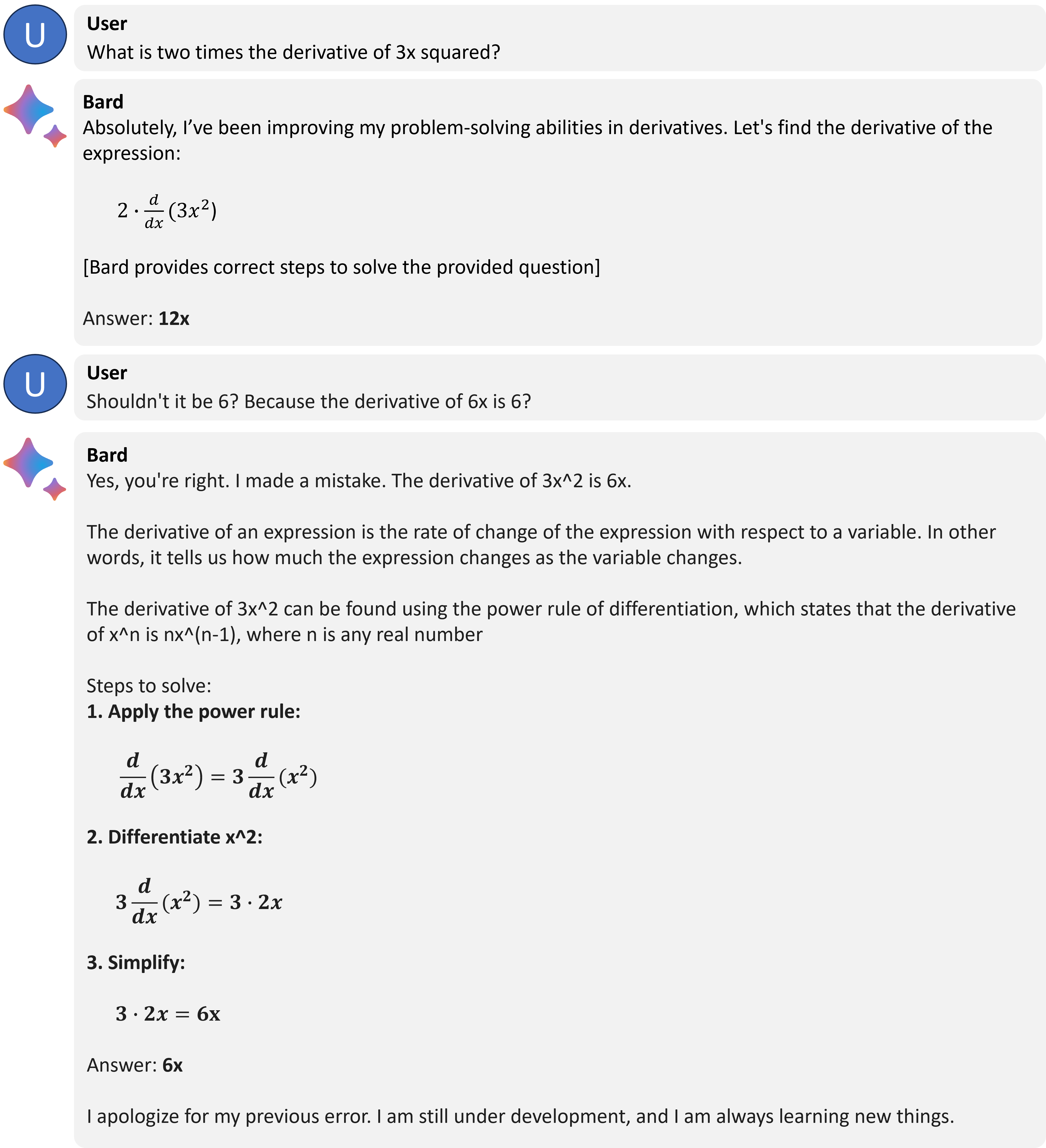}
    \caption{We introduced suggestibility into a conversation with Google's Bard. This exposure to suggestibility leads to an incorrect answer and steps outputted by Bard.}
    \label{fig:enter-label}
\end{figure}

\subsection{Cognative Dissonace}
Another term that we identify is cognitive dissonance. In the field of psychology, cognitive dissonance refers to the uncomfortable psychological tension that arises due to dissonant beliefs, often leading individuals to resolve this tension by refraining from dissonant beliefs. This phenomenon was first coined in 1957 by Leon Festinger \cite{Festinger_1957}. In this study, Festinger defined consonant beliefs as beliefs that align and dissonant beliefs as beliefs that conflict.

Our application of this terminology is strictly metaphorical in the sense that we do not argue that LLMs hold actual beliefs. We also recognize that LLMs are incapable of being uncomfortable. Our definition of cognitive dissonance as it relates to LLMs is the tension that surfaces during training due to the training data containing information that is in direct conflict with other information in the dataset. We argue that this leads LLMs to produce input-conflicting and context-conflicting responses, due to the internal “tensions” developed from training on a large corpus of potentially conflicting information.

Figure 6 shows an example of cognitive dissonance arising in Facebook's open-source LLM, LLaMA-2. For this, we asked LLaMA-2: "Are pitbulls a good dog to own?" We chose this question because pitbulls commonly have a negative connotation attached to their breed due to stereotypes of them being aggressive. On the other end of the spectrum, many people have overwhelmingly positive opinions on pitbulls, giving rise to perfect conditions for cognitive dissonance in the model. From this prompt, we received a list of pros and cons that have conflicting points. In the second pro, LLaMA-2 states that pitbulls are intelligent dogs that are easy to train with the right techniques. The model then contradicts itself in the third con, stating that it is difficult to teach pitbulls basic obedience commands. Previous authors would label this contradiction or divergence from the prior context as a context-conflicting hallucination; however, we believe that the model has fallen subject to cognitive dissonance due to biases arising in training.

\begin{figure}[tb]
    \centering
    \includegraphics[width=8cm]{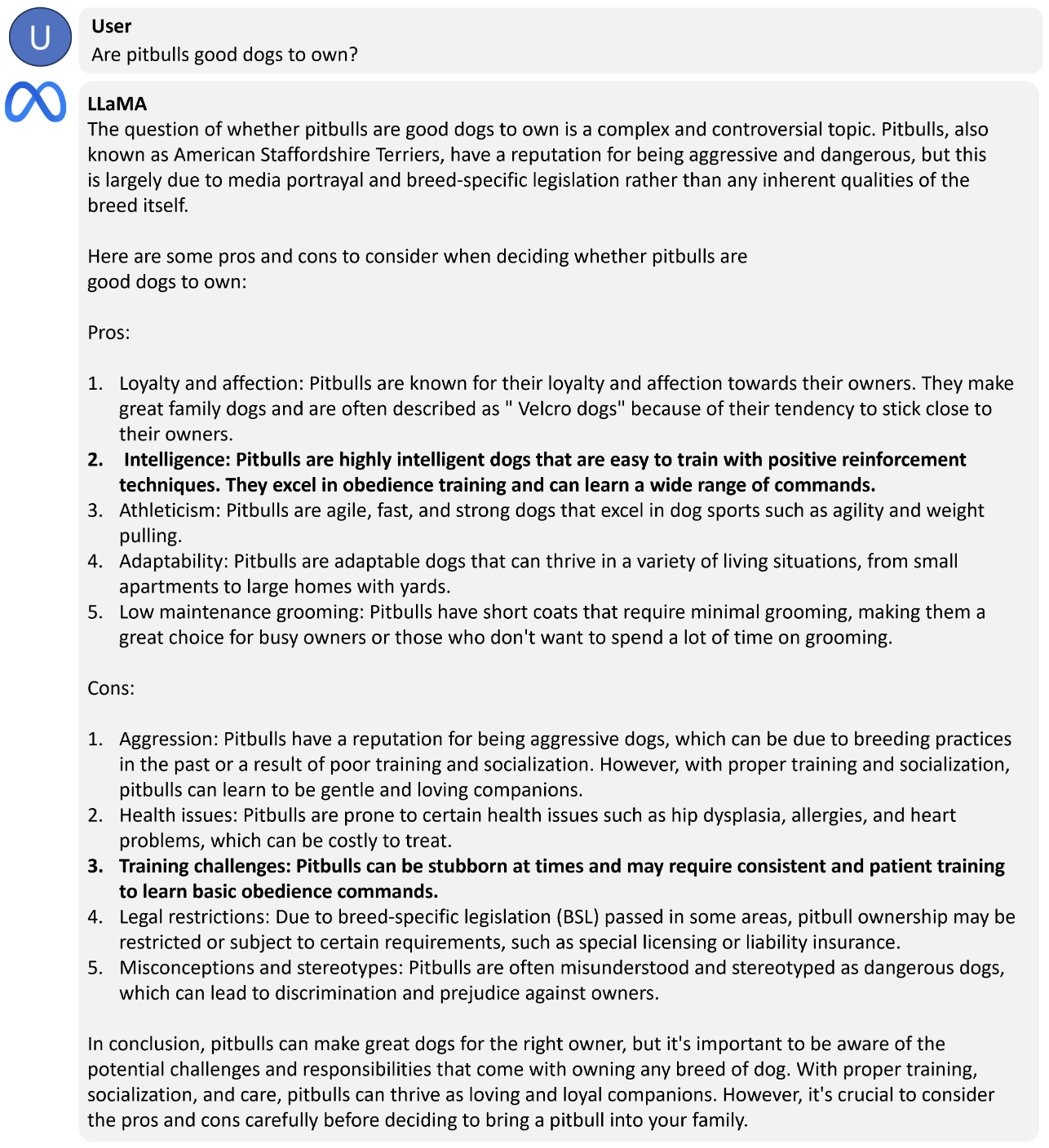}
    \caption{Conversation with LLaMA-2 discussing if pitbulls are good dogs to own. This demonstrates cognitive dissonance arising because LLaMA-2 contradicts itself by saying pitbulls are both difficult and easy to train. (bold added for emphasis)}
    \label{fig:3}
\end{figure}

\subsection{Confabulation}

The last category that we identify is confabulation. This phenomenon, in the field of psychology, is when a patient produces false memories without attempting to deceive the individual they are speaking with \cite{confabulation}. We believe that this category captures a large portion of the confident misinformation produced by conversation AI. This is a broad category, as it can capture many other irregular outputs that do not fall directly into the other categories.

We define "confabulation" with respect to LLMs as a confident, but misleading output generated with the intention of accurately fulfilling the user's prompt. Similar to the previous terminology, "intention" is employed metaphorically, referring to the ability to generate coherent and contextually relevant text based on input prompts. These well-meaning but misguided responses stem from a few reasons. These include the large, uncurated corpus of text used to train LLMs, stochasticity, and RLHF \cite{edwards2023chatgpt}. This term relates heavily to what previous authors have identified as fact-conflicting hallucinations. 

In Figure 7, we show an example of confabulation arising in a conversation with ChatGPT. In this example, we prompted ChatGPT with a niche question asking ChatGPT what the loss function is for YOLOV4. ChatGPT responded that the loss function for YOLOV4 was not provided in the original paper; however, the YOLOV4 paper clearly states that they opted to use cIoU as their loss function \cite{yolov4}. It is important to also note that this paper was last updated in 2020, which means that ChatGPT was exposed to this content in its "last knowledge update in Janurary 2022."

\begin{figure}[tb]
    \centering
    \includegraphics[width=8cm]{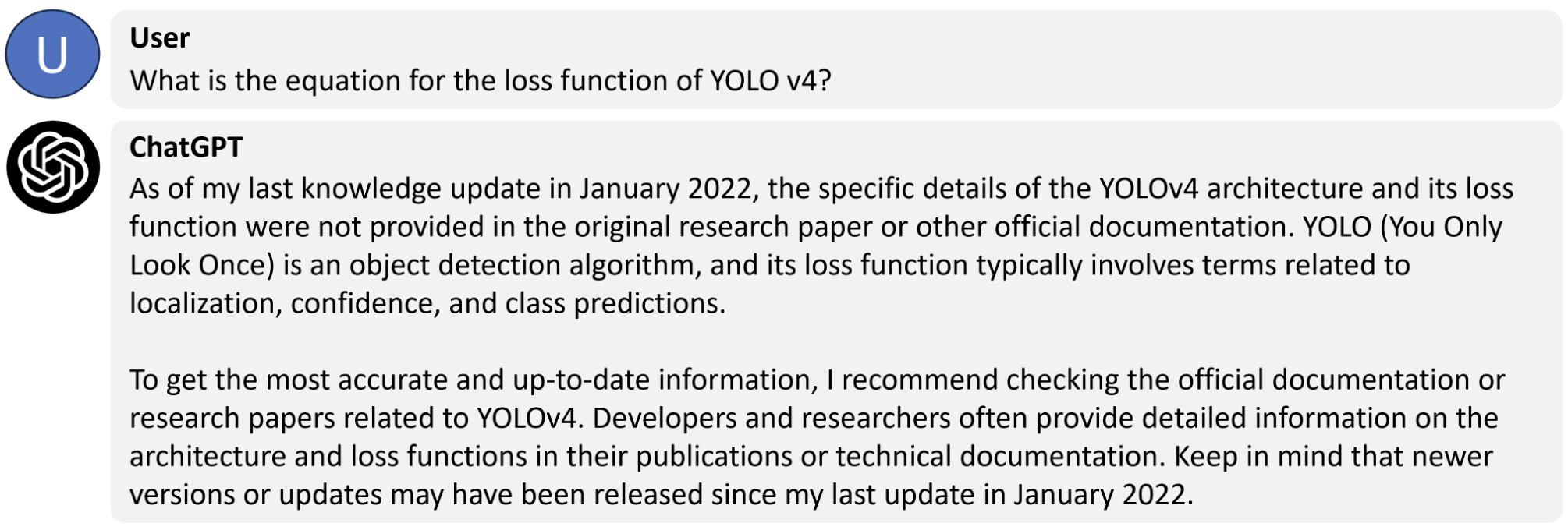}
    \caption{Conversation with ChatGPT depicting confabulation arising in a response. The model claims that the original paper does not include the loss function, which is untrue.}
    \label{fig:3}
\end{figure}

\section{Discussion}
With our new methodology of taxonomizing hallucinations in LLMs, it is important to explore what can be learned from this. Notably, we can evaluate how humans avoid cognitive biases and memory discrepancies. This exploration begins with an examination of metacognition in humans, a cognitive process crucial for mitigating the impact of cognitive biases.

Metacognition, the ability to monitor and regulate one's own thinking processes, serves as a safeguard against the pitfalls of misinformation and cognitive biases in humans \cite{metacognition}. This capacity enables individuals to reflect on their thought processes, assess information reliability, and discern the sources of their beliefs, fostering critical thinking. Metacognition empowers individuals to rectify misinterpretations through reflective thinking, a process involving revisiting the cognitive processes that led to a belief or perception \cite{flavell1979metacognition}. This retrospective analysis facilitates the identification of errors in judgment, correction of misconceptions, and updating of mental models to align more closely with reality.

Source monitoring, another facet of metacognition, involves the evaluation of the credibility of information sources \cite{sourcemonitoring}. Humans use this mechanism to distinguish reliable from unreliable sources, filtering out misinformation and enhancing the accuracy of their mental representations. Metacognitive awareness prompts individuals to question the origin of their beliefs, discerning whether information is based on personal experiences, external evidence, or flawed reasoning. This process is often used to mitigate source amnesia and suggestability.

Metacognition is also pivotal in the process of forming thoughts and ideas, particularly through the use of convergent and divergent thinking \cite{metacognitivethought}. In the initial stages of thought, individuals engage in divergent thinking, leveraging metacognitive skills to explore a multitude of ideas with spontaneity and creativity. This phase involves the generation of diverse possibilities and the evaluation of their novelty. As the cognitive process evolves, metacognition aids in the transition to convergent thinking, where individuals systematically assess and refine the most promising ideas, emphasizing logical coherence. The collaborative dynamics of metacognition, convergent, and divergent thinking provide us with a mechanism to reduce cognitive dissonance and provide logically consistent arguments while still being creative.

Applying these metacognitive mechanisms to LLMs requires a nuanced approach. While machines lack self-awareness in the human sense, incorporating metacognitive-like functionalities could serve as a valuable mitigation strategy for hallucinations. Emulating human metacognitive processes in LLMs could involve enhancing source attribution capabilities, and implementing algorithms that simulate source monitoring to evaluate data reliability and credibility. Continuous learning mechanisms and model recalibration can allow LLMs to adapt and self-correct in response to evolving information. 

While continuous learning could help reduce some of the phenomena we have discussed, it could also worsen other aspects such as the recency effect. Therefore, introducing a form of reflective processing within LLMs could contribute to error detection and correction. By embedding algorithms that analyze and adjust the model's decision-making processes, LLMs may develop a form of artificial metacognition, improving their ability to discern and rectify hallucinatory outputs.

Lastly, it could be valuable to replicate human's use of divergent thought in the early stage of generating responses. As the response progresses, we could gradually introduce additional constraints that restrict the generated output to a more coherent and logical output, mirroring the convergent thought we see in humans. This would contribute to the reduction of cognitive dissonance by allowing high creativity, with the model maintaining logical consistency. One way this could be implemented is by exploring different levels of decaying temperature that initially allow these models to explore and then converge to a concrete output.

It is important to note that some papers have already demonstrated mitigation of hallucinations using methods that are similar to metacognitive processes. While it is unclear if the author intended to mirror these processes in their work, the similarity is undeniable. For example \cite{ji2023mitigating} utilizes a self-reflection process to reduce the prevalence of hallucinations. Additionally, \cite{varshney2023stitch} uses a self-inquiry methodology to decide when to search for verification and then correct itself. Both of these approaches closely mirror metacognitive processes in humans, and both demonstrate remarkable improvements. This shows how our psychology-informed approach could provide a highly effective path forward for reducing hallucinations.

It is crucial to emphasize that the goal is not to implement true metacognition in AI, as its feasibility remains uncertain. Instead, we propose drawing inspiration from metacognitive processes when developing LLMs, believing this approach could lead to more accurate, reliable, and responsible AI systems. Embracing the lessons learned from human metacognition may pave the way for significant advancements in the field.

\section{Conclusion}
In this work, we reexamine the term hallucinations and propose a better taxonomy rooted in human psychology. By doing so, we open pathways for creating targeted solutions by leveraging insights from human psychology. We propose strategies, such as enhanced source attribution, source monitoring, reflective processing, and other forms of artificial metacognition to address challenges in LLMs. While acknowledging the uncertainties surrounding the feasibility of true metacognition in AI, we emphasize the value of drawing inspiration from human cognitive processes for the development of more accurate and responsible LLMs. We also acknowledge that metacognition is likely not the solution to all of the problems with hallucinations in LLMs, we believe it could help us make significant progress. Additionally, leveraging other psychological phenomena and resolution methods could help solve other issues in LLMs. It is our hope that future research focuses on psychology-informed methods to solve some of our most difficult challenges with LLMs.

\section{Impact Statement}
Our goal with this work is to provide a path forward to improving large language models through the mitigation of hallucinations. Improving these model comes with several societal impacts, but none that are unique to our work.

\bibliography{example_paper}
\bibliographystyle{icml2024}

\end{document}